\documentclass{bmvc2k}
\usepackage{bbm}
\usepackage{amsthm}
\usepackage{xcolor}
\usepackage{caption}

\newtheorem*{notation}{Notation}
\newtheorem*{defn}{Definition}
\newtheorem{prop}{Proposition}

\title{Learning Neural Network Architectures using Backpropagation}

\addauthor{Suraj Srinivas}{surajsrinivas@grads.cds.iisc.ac.in}{1}
\addauthor{R. Venkatesh Babu}{venky@cds.iisc.ac.in}{1}

\addinstitution{
 Department of Computational \\and Data Sciences\\
 Indian Institute of Science\\
 Bangalore, India 
}

\runninghead{Srinivas and Babu}{Learning Neural Architectures}

\begin{document}

\maketitle

\begin{abstract}
Deep neural networks with millions of parameters are at the heart of many state of the art machine learning models today. However, recent works have shown that models with much smaller number of parameters can also perform just as well. In this work, we introduce the problem of architecture-learning, i.e; learning the architecture of a neural network along with weights. We start with a large neural network, and then learn which neurons to prune. To this end, we introduce a new trainable parameter called the \emph{Tri-State ReLU}, which helps in pruning unnecessary neurons. We also propose a smooth regularizer which encourages the total number of neurons after elimination to be small. The resulting objective is differentiable and simple to optimize. We experimentally validate our method on both small and large networks, and show that it can learn models with considerably smaller number of parameters without affecting prediction accuracy.
\end{abstract}

\section{Introduction}

\begin{quote}
\textit{Everything should be made as simple as possible, but not simpler} -~ Einstein
\end{quote}
For large-scale tasks like image classification, the general practice in recent times has been to train large networks with many millions of parameters (see \cite{krizhevsky2012imagenet, Simonyan15, Szegedy_2015_CVPR}). Looking at these models, it is natural to ask - are so many parameters really needed for good performance? In other words, are these models as simple as they can be? A smaller model has the advantage of being faster to evaluate and easier to store - both of which are crucial for real-time and embedded applications. In this work, we consider the problem of automatically building smaller networks that achieve performance levels similar to larger networks. 

Regularizers are often used to encourage learning simpler models. These usually restrict the magnitude ($\ell_2$) or the sparsity ($\ell_1$) of weights. However, to restrict the computational complexity of neural networks, we need a regularizer which restricts the width and depth of network. Here, \emph{width} of a layer refers to the number of neurons in that layer, while \emph{depth} simply corresponds to the total number of layers. Generally speaking, the greater the width and depth, the more are the number of neurons, the more computationally complex the model is. Naturally, one would want to restrict the total number of neurons as a means of controlling the computational complexity of the model. However, the number of neurons is an integer, making it difficult to optimize over. This work aims at making this problem easier to solve. 

The overall contributions of the paper are as follows.
\begin{itemize}
\item We propose novel trainable parameters which are used to restrict the total number of neurons in a neural network model - thus effectively selecting width and depth (Section 2)
\item We perform experimental analysis of our method to analyze the behaviour of our method. (Section 4)
\item We use our method to perform architecture selection and learn models with considerably small number of parameters (Section 4)
\end{itemize}

\section{Complexity as a regularizer}
In general, the term `architecture' of a neural network can refer to aspects of a network other than width and depth (like filter size, stride, etc). However, here we use that word to simply mean width and depth. Given that we want to reduce the complexity of the model, let us formally define our notions of complexity and architecture.

\begin{notation}
 Let $\Phi = [n_1,n_2,...,n_m,0,0,...]$ be an infinite-dimensional vector whose first $m$ components are positive integers, while the rest are zeros. This represents an $m$-layer neural network architecture with $n_{i}$ neurons for the $i^{th}$ layer. We call $\Phi$ as the architecture of a neural network.
\end{notation}

For these vectors, we define an associated norm which corresponds to our notion of architectural complexity of the neural network. Our notion of complexity is simply the total number of neurons in the network. 

The true measure of computational complexity of a neural network would be the total number of weights or parameters. However, if we consider a single layer neural network, this is proportional to the number of neurons in the hidden layer. Even though this equivalence breaks down for multi-layered neural networks, we nevertheless use the same for want of simplicity.

\begin{defn}
The complexity of a $m$-layer neural network with architecture $\Phi$ is given by $\| \Phi \| = \sum\limits_{i=1}^{m} n_i $. 
\label{defn:complexity}
\end{defn}

Our overall objective can hence be stated as the following optimization problem.

\begin{equation}
\hat{\theta}, \hat{\Phi} = \underset{\theta,\Phi}{\arg\min} ~\ell(\hat{y}(\theta, \Phi),y) + \lambda \| \Phi \| 
\label{eqn:main}
\end{equation}
where $\theta$ denotes the weights of the neural network, and $\Phi$ the architecture. $\ell(\hat{y}(\theta,\Phi),y)$ denotes the loss function, which depends on the underlying task to be solved. For example, squared-error loss functions are generally used for regression problems and cross-entropy loss for classification. In this objective, there exists the classical trade-off between model complexity and loss, which is handled by the $\lambda$ parameter. Note that we learn both the weights ($\theta$) as well as the architecture ($\Phi$) in this problem. We term any algorithm which solves the above problem as an \emph{Architecture-Learning (AL)} algorithm. 

We observe that the task defined above is very difficult to solve, primarily because $\|\Phi\|$ is an integer. This makes it an integer programming problem. Hence, we cannot use gradient-based techniques to optimize for this. The main contribution of this work is the re-formulation of this optimization problem so that Stochastic Gradient Descent (SGD) and back-propagation may be used. 

\subsection{A Strategy for a trainable regularizer}
\begin{figure}[h]
\begin{tabular}{c  c}
\includegraphics[width=6cm,trim={0cm 8cm 4.5cm 0cm},clip]{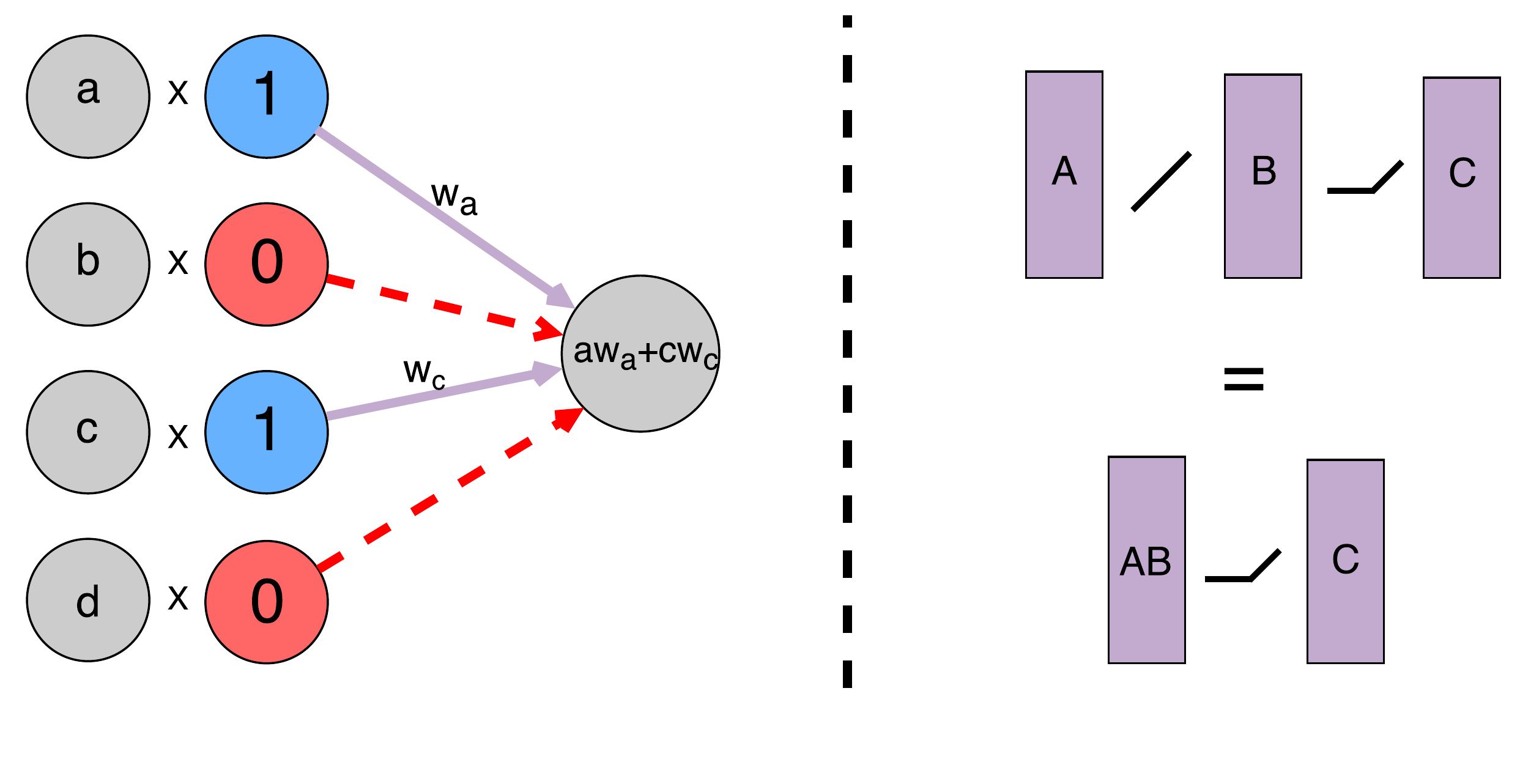}&
 \hspace{2cm}\includegraphics[width=4cm]{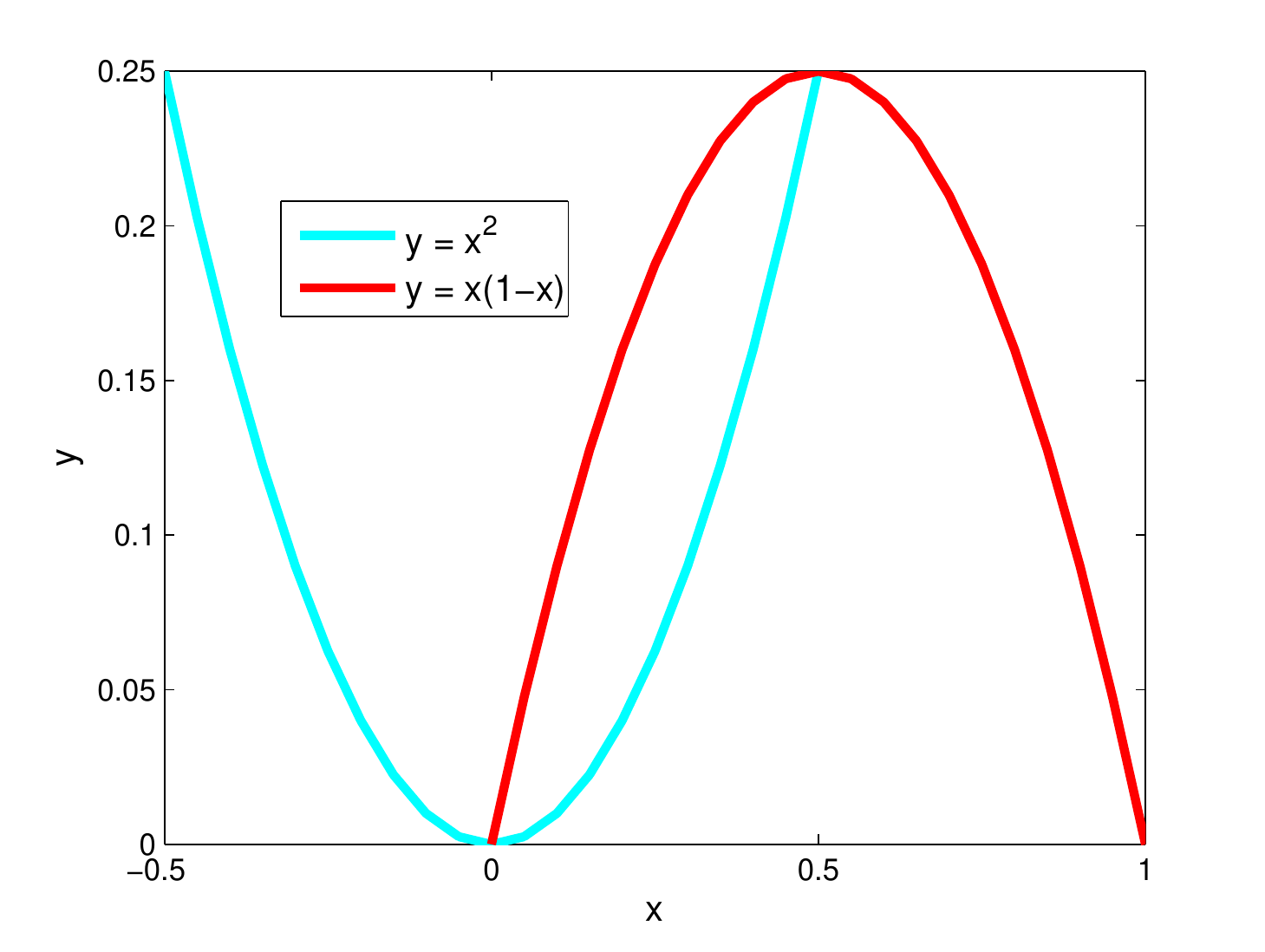}\\
(a)&\hspace{2cm}(b)
\end{tabular}
\caption{\small{\textbf{(a)} Our strategy for selecting width and depth. Left: Grey blobs denote neurons, coloured blobs denote the proposed additional trainable parameters. Right: Purple bars denote weight-matrices. 
\\\textbf{(b)} Graph of the $\ell_2$ regularizer and the binarizing regularizer in 1-D.}}
\label{fig:NN}
\end{figure}

We require a strategy to automatically select a neural network's architecture, i.e; the width of each layer and depth of the network. One way to select for width of a layer is to introduce additional learnable parameters which multiply with every neuron's output, as shown in Figure \ref{fig:NN}\textcolor{red}{(a)}. If these new parameters are restricted to be binary, then those neurons with a zero-parameter can simply be removed. In the figure, the trainable parameters corresponding to neurons with values $b$ and $d$ are zero, nullifying their contribution. Thus, the sum of these binary trainable parameters will be equal to the effective width of the network. For convolutional layers with $n$ feature map outputs, we have $n$ additional parameters that select a subset of the $n$ feature maps. A single additional parameter multiplies with an entire feature map either making it zero or preserving it. After all, filters are analogous to neurons for convolutional layers.

To further reduce the complexity of network, we also strive to reduce the network's depth. It is well known that two neural network layers without any non-linearity between them is equivalent to a single layer, whose parameters are given by the matrix product of the weight matrices of the original two layers. This is shown on the right of Figure \ref{fig:NN}\textcolor{red}{(a)}. We can therefore consider a trainable non-linearity, which prefers `linearity' over `non-linearity'. Wherever linearity is selected, the corresponding layer can be combined with the next layer. Hence, the total complexity of the neural network would be the number of parameters in layers with a non-linearity.
 
In this work, we combine both these intuitive observations into one single framework. This is captured in our definition of the \emph{tri-state ReLU} which follows. 

\subsubsection{Definition: Tri-state ReLU}
We define a new trainable non-linearity which we call the tri-state ReLU (tsReLU) as follows:
\begin{equation}
tsReLU(x) =
\begin{cases}
wx, & x \geq 0 \\
wdx, & otherwise
\end{cases}
\end{equation}

This reduces to the usual ReLU for $w = 1$ and $d = 0$. For a fixed $w = 1$ and a trainable $d$, this turns into parametric ReLU \cite{he2015delving}. For us, both $w$ and $d$ are trainable. However, we restrict both these parameters to take only binary values. As a result, three possible states exist for this function. For $w = 0$, this function is always returns zero. For $w = 1$ and $d = 0$ it behaves similar to ReLU, while for $w = d = 1$ it reduces to the identity function. 

Here, parameter $w$ selects for the \textbf{width} of the layer, while $d$ decides \textbf{depth}. While the $w$ parameter is different across channels of a layer, the $d$ parameter is tied to the same value across all channels. If $d = 1$, we can combine that layer with the next to yield a single layer. If $w = 0$ for any channel, we can simply remove that neuron as well as the corresponding weights in the next layer.

Thus, our objective while using the tri-state ReLU is


\begin{equation}
\begin{aligned}
& \underset{\theta,w_{ij},d_i:\forall i,j}{\text{Minimize}} 
& & \ell(\hat{y}(\theta,\mathbf{w},\mathbf{d}),y) \\
& \text{such  that} 
& & w_{ij},d_i \in \{0,1\} \\
&&& \forall i,j
\end{aligned}
\end{equation}

We remind the reader that here $i$ denotes the layer number, while $j$ denotes the $j^{th}$ neuron in a layer. Note that for $\lambda = 0$, it converts the objective in Equation \ref{eqn:main} from an integer programming problem to that of binary programming.

\subsubsection{Learning binary parameters}
Given the definition of tri-state ReLU (tsReLU) above, we require a method to learn binary parameters for $w$ and $d$. To this end, we use a regularizer given by $w \times (1-w)$ \cite{murray2010algorithm}. This regularizer encourages binary values for parameters, if they are constrained to lie in $[0,1]$. 

Henceforth, we shall refer to this as the \textit{binarizing} regularizer. Murray and Ng \cite{murray2010algorithm} showed that this regularizer does indeed converge to binary values given a large number of iterations.  For the 1-D case, this function is an downward-facing parabola with minima at $0$ and $1$, as shown in Figure \ref{fig:NN}\textcolor{red}{(b)}. As a result, weights ``fall'' to $0$ or $1$ at convergence. In contrast, the $\ell_2$ regularizer is an upward facing parabola with a minimum at $0$, which causes it to push weights to be close to zero. 

With this intuition, we now state our tsReLU optimization objective.

\begin{multline}
\theta, \textbf{w}, \textbf{d} = \underset{\theta,w_{ij},d_i:\forall i,j}{\arg\min} ~\ell(\hat{y}(\theta,\textbf{w},\textbf{d}),y) + \lambda_1 \sum\limits_{i=1}^{m} \sum\limits_{j=1}^{n_i} w_{ij}(1-w_{ij}) + \lambda_2 \sum\limits_{i=1}^{m}d_i(1-d_i)
\label{eqn:objective}
\end{multline}

Note that $\lambda_1$ is the regularization constant for the width-limiting term, while $\lambda_2$ is for the depth-limiting term. This objective can be solved using the usual back-propagation algorithm. As indicated earlier, this binarizing regularizer works only if $w$'s and $d$'s are guaranteed to be in $[0,1]$. To enforce the same, we perform clipping after parameter update. 


 
 


After optimization, even though the final parameters are expected to be close to binary, they are still real numbers close to $0$ or $1$. Let $w_{ij}$ be the parameter obtained during the optimization. The tsReLU function uses a binarized version of this variable 

\begin{equation*}
w'_{ij} = \begin{cases}
1, & w_{ij} \geq 0.5 \\
0, & otherwise
\end{cases}
\end{equation*}

during the feedforward stage. Note that $w_{ij}$ slowly changes during training, while $w'_{ij}$ only reflects the changes made to $w_{ij}$. A similar equation holds for $d'_i$.

\subsection{Adding model complexity}
So far, we have considered the problem of solving Equation \ref{eqn:main} with $\lambda = 0$. As a result, the objective function described above does not necessarily select for smaller models. Let $h_i =  \sum\limits_{j=1}^{n_i} w_{ij}$ correspond to the complexity of a layer. The model complexity term is given by 

\begin{equation*}
\| \Phi \| = \sum\limits_{i=1}^{m} h_i   ~\mathbbm{1}_{(d_i = 0)}
\end{equation*}

This is formulated such that for $d_i = 0$, the complexity in a layer is just $h_i h_{i+1}$, while for $d_i = 1$ (non-linearity absent), the complexity is $0$. Overall, it counts the total number of weights in the model at convergence.

We now add a regularizer analogous to model complexity (defined above) in our optimization objective in Equation \ref{eqn:objective} . Let us call the regularizer corresponding to model complexity as $R_m(\textbf{h},\textbf{d})$, which is given by

\begin{equation}
R_m(\textbf{h},\textbf{d}) = \lambda_3 \sum\limits_{i=1}^{m} h_i \mathbbm{1}_{(d_i < 0.5)}  - \lambda_4 \sum\limits_{i=1}^{m} d_{i}
\label{eqn:model_complexity}
\end{equation}

The first term in the above equation limits the complexity of each layer's width, while the second term limits the network's depth by encouraging linearity. Note that the first term becomes zero when a non-linearity is absent. Also note that the indicator function in the first term is non-differentiable. As a result, we simply treat that term as a constant with respect to $d_i$.

\section{Related Work}
There have been many works which look at performing compression of a neural network. Weight-pruning techniques were popularized by \emph{Optimal Brain Damage} \cite{lecun1989optimal} and \emph{Optimal Brain Surgery} \cite{hassibi1993second}. Recently, \cite{BMVC2015_31} proposed a neuron pruning technique, which relied on neuronal similarity. Our work, on the other hand, performs neuron pruning based on learning, rather than hand-crafted rules. Our learning objective can thus be seen as performing pruning and learning together, unlike the work of Han \emph{et al.} \cite{han2015learning}, who perform both operations alternately. 

Learning neural network architecture has also been explored to some extent. The \emph{Cascade-correlation} \cite{fahlman1989cascade} proposed a novel learning rule to `grow' the neural network. However, it was shown for only a single layer network and is hence not clear how to scale to large deep networks. Our work is inspired from the recent work of Kulkarni \emph{et al.} \cite{BMVC2015_23} who proposed to learn the width of neural networks in a way similar to ours. Specifically, they proposed to learn a diagonal matrix $D$ along with neurons $Wx$, such that $DWx$ represents that layer's neurons. However, instead of imposing a binary constraint (like ours), they learn real-weights and impose an $\ell_1$-based sparsity-inducing regularizer on $D$ to encourage zeros. By imposing a binary constraint, we are able to directly regularize for the model complexity. Recently, Bayesian Optimization-based algorithms \cite{snoek2012practical} have also been proposed for automatically learning hyper-parameters of neural networks. However, for the purpose of selecting architecture, these typically require training multiple models with different architectures - while our method selects the architecture in a single run. A large number of evolutionary algorithms (see \cite{yao1999evolving, stanley2002evolving, stanley2009hypercube}) also exist for the task of finding Neural Network architectures. However, these are typically evaluated on small scale problems, often not relating to pattern recognition tasks.


Many methods have been proposed to train models that are deep, yet have a lower parameterisation than conventional networks. Collins and Kohli \cite{DBLP:journals/corr/CollinsK14} propose a sparsity inducing regulariser for backpropogation which promotes many weights to have zero magnitude. They achieve reduction in memory consumption when compared to traditionally trained models. In contrast, our method promotes \emph{neurons} to have a zero-magnitude. As a result, our overall objective function is much simpler to solve. Denil \emph{et al.} \cite{denil2013predicting} demonstrate that most of the parameters of a model can be \textit{predicted} given only a few parameters. At training time, they learn only a few parameters and predict the rest. Yang \emph{et al.} \cite{yang2014deep} propose an \emph{Adaptive Fastfood transform}, which is an efficient re-parametrization of fully-connected layer weights. This results in a reduction of complexity for weight storage and computation.

Some recent works have also focussed on using approximations of weight matrices to perform compression. Jaderberg \emph{et al.} \cite{jaderberg2014speeding} and Denton \emph{et al.} \cite{denton2014exploiting} use SVD-based low rank approximations of the weight matrix. Gong \emph{et al.} \cite{gong2014compressing} use a clustering-based product quantization approach to build an indexing scheme that reduces the space occupied by the matrix on disk. 

\section{Experiments}
In this section, we perform experiments to analyse the behaviour of our method. In the first set of experiments, we evaluate performance on the MNIST dataset. Later, we look at a case study on ILSVRC 2012 dataset. Our experiments are performed using the Theano Deep Learning Framework \cite{bergstra2010theano}.

\subsection{Compression performance}
We evaluate our method on the MNIST dataset, using a LeNet-like \cite{lecun1998gradient} architecture. The network consists of two $5 \times 5$ convolutional layers with 20 and 50 filters, and two fully connected layers with 500 and 10 (output layer) neurons. We use this architecture as a starting point to learn smaller architectures. First, we learn using our additional parameters and regularizers. Second, we remove neurons with zero gate values and collapse depth for linearities wherever is it advantageous. For example, it might not be advantageous to remove depth in a bottleneck layer (like in auto-encoders). Thus, the second part of the process is human-guided.

Starting from a baseline architecture, we learn smaller architectures with variations of our method. Note that there is max-pooling applied after each of the convolutional layers, which rules out depth selection for those two layers. We compare the proposed method against baselines of directly training a neural network (NN) on the final architecture, and our method of learning a fixed final width (FFW) for various layers. In Table \ref{table: mnist-archlearn}, the \emph{Layers Learnt} column has binary elements $(w,d)$ which denotes whether width($w$) or depth($d$) are learnt for each layer in the baseline network. As an example, the second row shows a method where only the \emph{width} is learnt in the first two layers, and \emph{depth} also learnt in the third layer. This table shows that all considered models - large and small - perform more or less equally well in terms of accuracy. This empirically shows that the small models discovered by AL preserve accuracy.

We also compare the compression performance of our AL method against SVD-based compression of the weight matrix in Table \ref{table: mnist-compression}. Here we only compress layer 3 (which has $800\times 500$ weights) using SVD. The results show that learning a smaller network is beneficial over learning a large network and then performing SVD-based compression. 

\begin{table*}
\centering
\begin{tabular}{c|c|c|c|c||c}
\hline 
\textbf{Method} & \textbf{$\lambda_3$}& \textbf{Layers Learnt} & \textbf{Architecture} & \textbf{AL (\%)} & \textbf{NN (\%)} \\  
\hline \hline
Baseline & N/A & (0,x)-(0,x)-(0,0) & 20-50-500-10 & N/A & 99.3  \\
AL$_1$ & $0.4  \lambda_1$ &(1,x)-(1,x)-(1,1) & 16-26-10 & 99.07 & 99.08 \\
AL$_2$ & $0.4  \lambda_1$ &(1,x)-(1,x)-(1,0) & 20-50-20-10 & 99.07 & 99.14  \\
AL$_3$ & $0.2  \lambda_1$ &(1,x)-(1,x)-(1,1) & 16-40-10 & 99.22 & 99.25  \\
AL$_4$ & $0.2  \lambda_1$ &(1,x)-(1,x)-(1,0) & 20-50-70-10 & 99.19 & 99.21 \\

\hline 
\end{tabular} 
\vspace{5pt}
\caption{\small{Architecture learning performance of our method on a LeNet-like baseline. The \emph{Layers Learnt} column has binary elements $(w,d)$ which denotes whether width($w$) or depth($d$) are learnt for each layer in the baseline network. 
AL = Architecture Learning, NN = Neural Network trained w/o AL}}
\label{table: mnist-archlearn}
\vspace{-0.5cm}
\end{table*}

\begin{table}[htbp]
\centering
\begin{tabular}{c|c|c}
\hline 
\textbf{Method} & \textbf{Params} & \textbf{Accuracy (\%)} \\  
\hline \hline
Baseline & 431K & 99.3 \\
SVD (rank-10) & 43.6K & 98.47 \\
AL$_2$ & \textbf{40.9K} & \textbf{99.07} \\
SVD (rank-40) & 83.1K & 99.06 \\
AL$_4$ & \textbf{82.3K} & \textbf{99.19} \\


\hline 
\end{tabular} 
\vspace{5pt}
\caption{\small{Comparison of compression performance of proposed method against SVD-based weight-matrix compression.}}
\label{table: mnist-compression}
\vspace{-0.5cm}
\end{table}



\subsection{Analysis}
We now perform a few more experiments to further analyse the behaviour of our method. In all cases, we train `AL$_2$'-like models, and consider the third layer for evaluation. We start learning with the baseline architecture considered above.

\begin{figure}[h]
\begin{tabular}{cc}
\includegraphics[width=6cm]{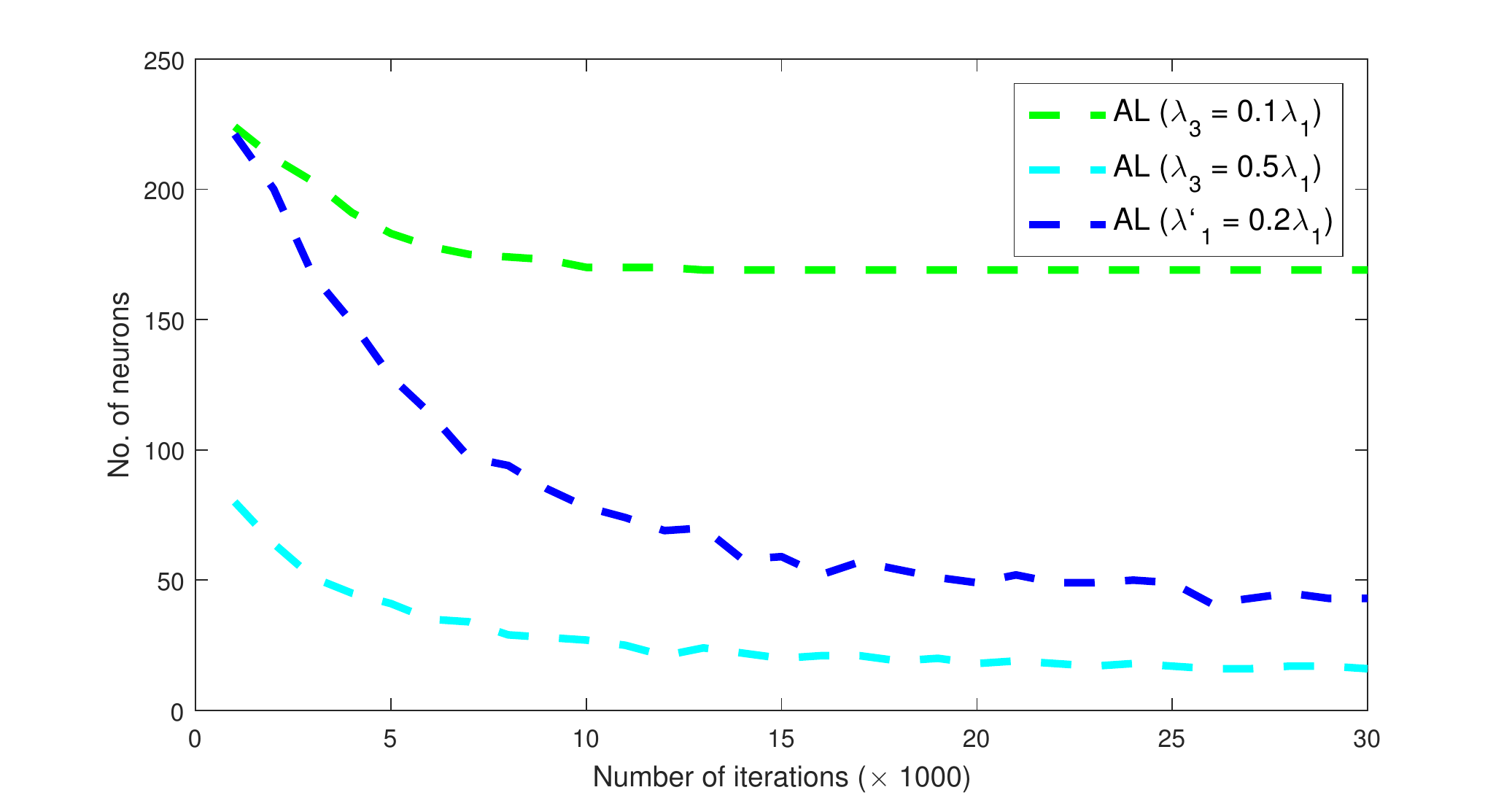}&
\includegraphics[width=6cm]{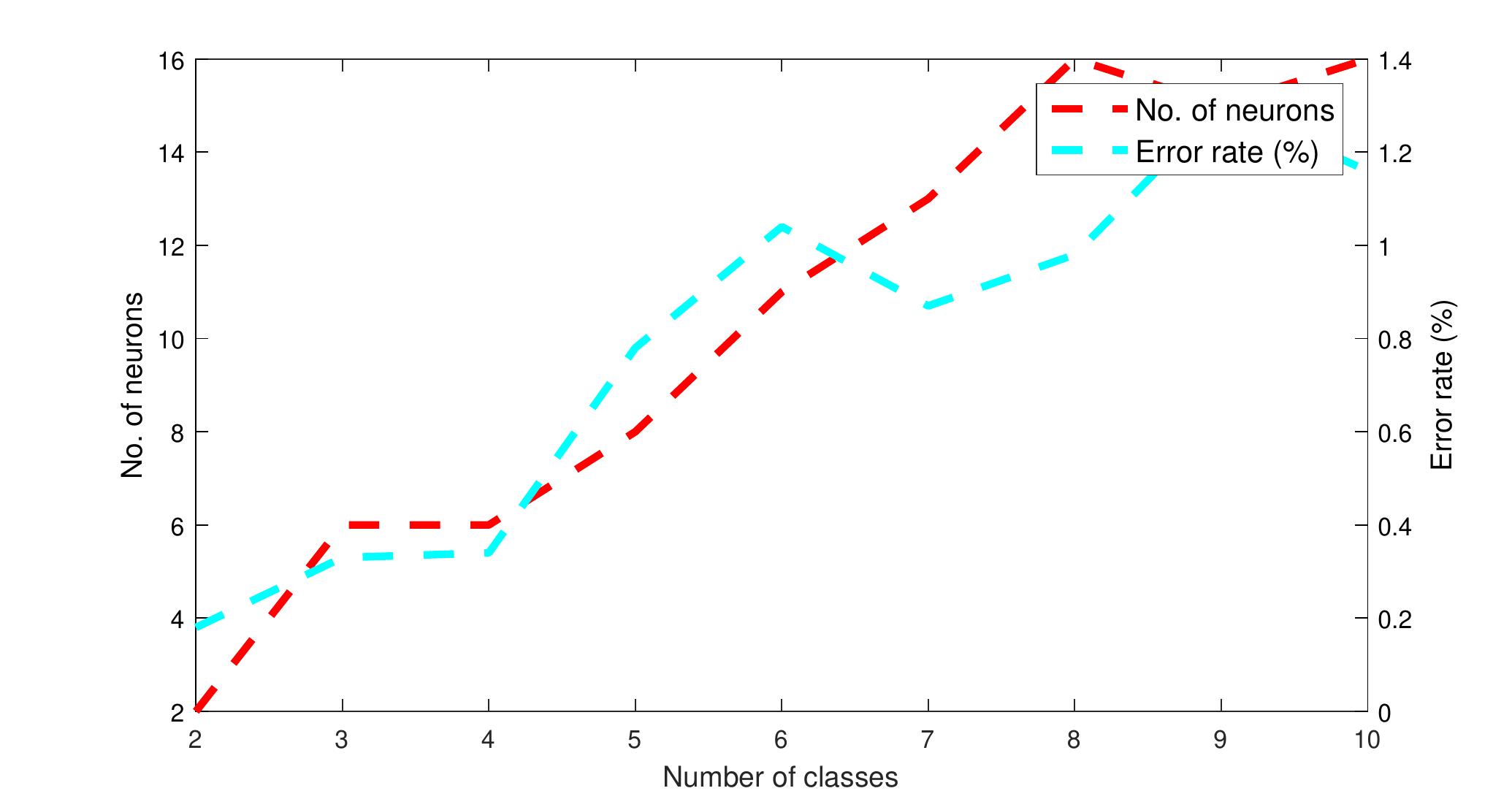}\\
(a)&(b)
\end{tabular}
\caption{\small{\textbf{(a)} Plot of the architecture learnt against the number of iterations. We see that $\lambda_1$ affects convergence rate while $\lambda_3$ affects amount of pruning.
\\\textbf{(b)} Plot of the no. of neurons learnt for MNIST with various number of classes. We see that both the neuron count and the error rate increase with increase in number of classes.}}
\label{fig:dynamics}
\vspace{-0.5cm}
\end{figure}

First, we look at the effects of using different hyper-parameters . From Figure \ref{fig:dynamics}\textcolor{red}{(a)} , we observe that (i) increasing $\lambda_3$ encourages the method to prune more, and (ii) decreasing $\lambda_1$ encourages the method to learn the architecture for an extended amount of time. In both cases, we see that the architecture stays more-or-less constant after a large enough number of iterations.

Second, we look at the learnt architectures for different amounts of data complexity. Intuitively, simpler data should lead to smaller architectures. A simple way to obtain data of differing complexity is to simply vary the number of classes in a multi-class problem like MNIST. We hence vary the number of classes from $2 - 10$, and run our method for each case without changing any hyper-parameters. As seen in Figure \ref{fig:dynamics}\textcolor{red}{(b)} , we see an almost monotonic increase in both architectural complexity and error rate, which confirms our hypothesis.

Third, we look at the depth-selection capabilities of our method. We used models with various initial depths and observed the depths of the resultant models. We used an initial architecture of 20 - 50 - (75 $\times$ n) - 10, where layers with width 75 are repeated to obtain a network of desired depth. We see that for small changes in the initial depth, the final learnt depth stays more or less constant. \footnote{For theoretical analysis of our method, please see the Supplementary material.}

%

\begin{table}[htbp]
\centering
\begin{tabular}{c|c|c|c}
\hline 
\textbf{Initial} & \textbf{Final} &\textbf{Learnt Architecture} & \textbf{Error (\%)} \\  
\textbf{Depth} & \textbf{Depth} &\textbf{} & \textbf{} \\  
\hline \hline
6 & 5 & 18-31-32-24-10 & 1.02 \\
8 & 6 & 17-37-39-29-21-10 & 0.99 \\
10 & 6 & 17-34-32-21-21-10 & 0.97 \\
12 & 6 & 18-34-30-21-17-10  & 1.04\\
15 & 8 & 16-37-35-25-20-22-10 & 0.93\\
\hline 
\end{tabular} 
\vspace{5pt}
\caption{\small{Performance of the proposed method on networks of increasing depth.}}
\label{table: mnist-depth}
\vspace{-0.5cm}
\end{table}

\subsection{Architecture Selection}
In recent times, Bayesian Optimization (BO) has emerged as a compelling option for hyper-parameter optimization. In these set of experiments, we compare the architecture-selection capabilities of our method against BO. In particular, we use the Spearmint-lite software package \cite{snoek2012practical} with default parameters for our experiments. 

\begin{figure}[h]
\begin{tabular}{cc}
\includegraphics[width=6cm,trim={1cm 1cm 1.5cm 0.5cm},clip]{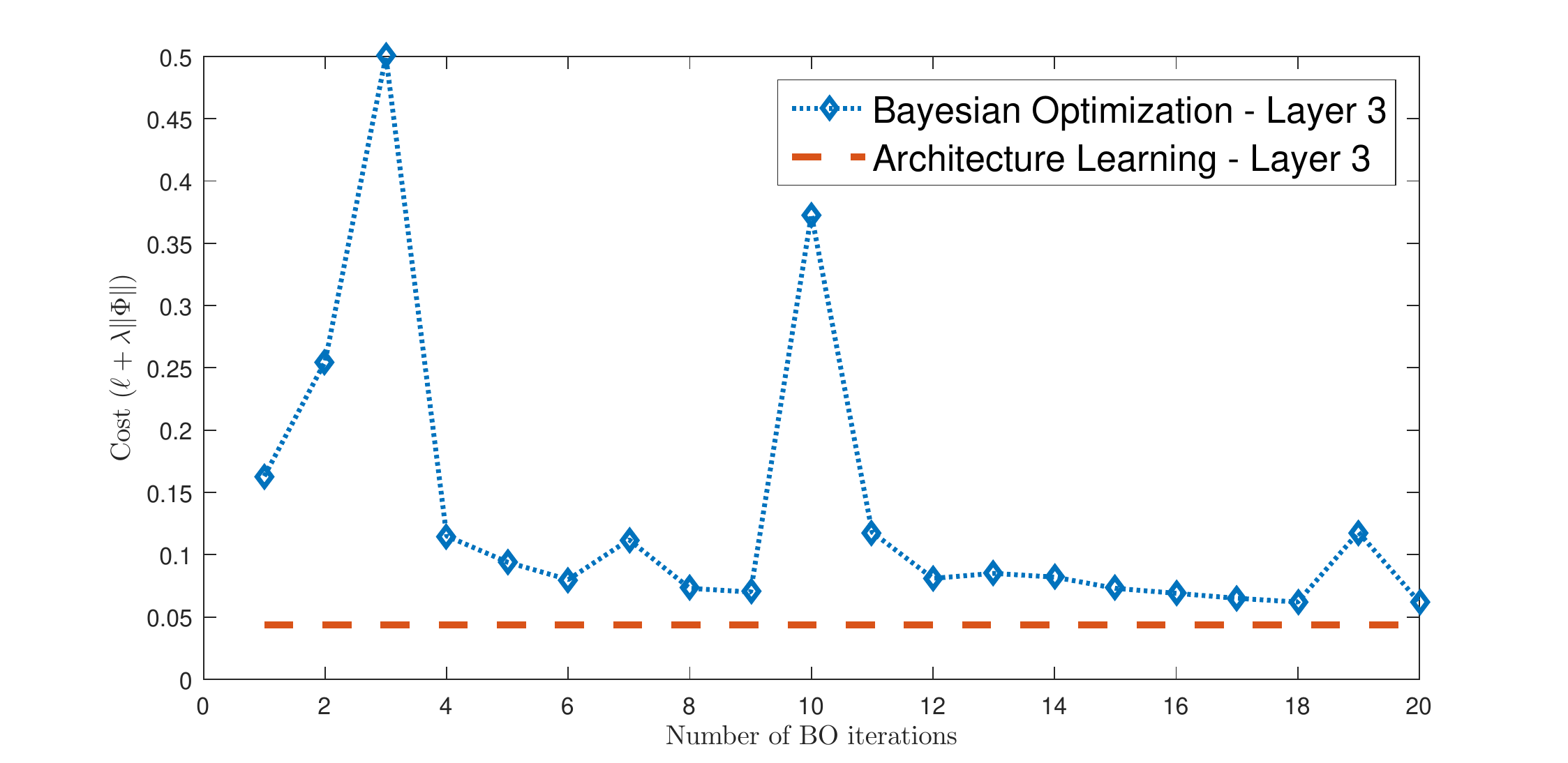}&
\includegraphics[width=6cm,trim={1cm 1cm 1.5cm 0.5cm},clip]{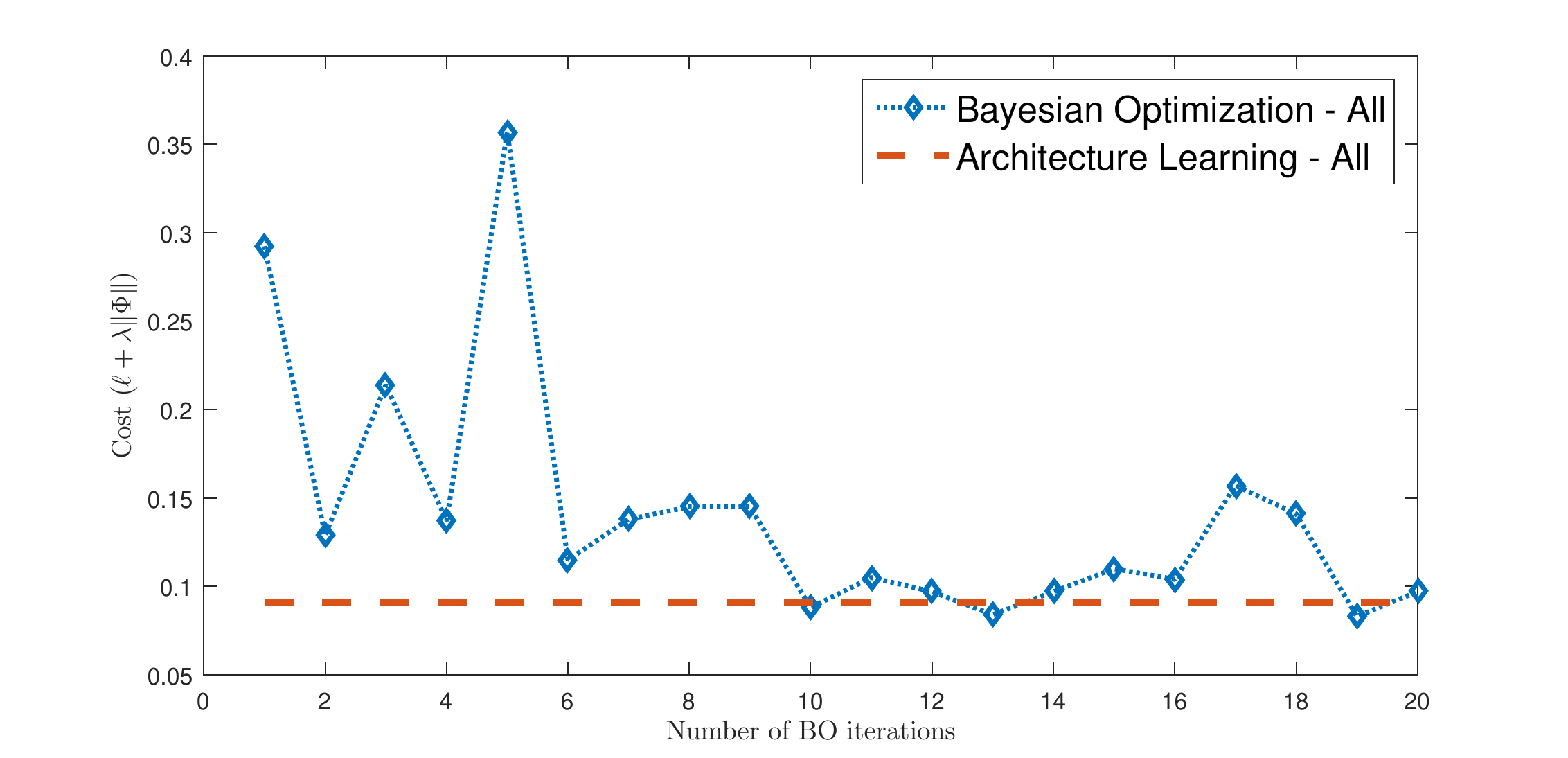}\\
(a)&(b)
\end{tabular}
\caption{\small{\textbf{(a)} Comparison against Bayesian Optimization for the case of learning the width of only third layer. 
\textbf{(b)} Similar comparison for learning the widths of all three layers. ($\lambda = 10^{-5}$ in both cases) }}
\label{fig:BO}
\vspace{-0.5cm}
\end{figure}

We use BO to first determine the width of the last FC layer (a single scalar), and later, the width of all three layers (3 scalars). For comparison, we use the same objective function for both BO and Architecture-Learning. This means that we use $\lambda_3 = 10^{-5}$ for AL, while we externally compute the cost after every training run for BO. Figure \ref{fig:BO} shows that BO typically needs multiple runs to discover networks which perform close to AL. Performing such multiple runs is often prohibitive for large networks. Even for a small network like ours, training took $\sim$30 minutes on a TitanX GPU for 300 epochs. Training with AL does not change the training time, whereas using BO we spent $\sim$10 hours for completing 20 runs. Further, AL directly optimizes the cost function as opposed to BO, which performs a black-box optimization. 

Given that we perform architecture selection, what hyper-parameters does AL need? We notice that we only need to decide four quantities - $\lambda_{1-4}$. If our objective is to only decide widths, we need to decide only two quantities - $\lambda_1$ and $\lambda_3$. Thus, for a $n$-layer neural network, we are able to decide $n$ (or $2n - 1$) numbers (widths and depths) based on only two (or four) global hyper-parameters. In the Appendix, we shall look at heuristics for setting these hyper-parameters.

\vspace{-0.25cm}
\subsection{Case study: AlexNet}
\vspace{-0.25cm}

For the experiments that follow, we use an AlexNet-like \cite{krizhevsky2012imagenet} model, called CaffeNet, provided with the Caffe Deep Learning framework. It is very similar to AlexNet, except that the order of max-pooling and normalization have been interchanged. We use the ILSVRC 2012 \cite{ILSVRC15} validation set to compute accuracies in the Table \ref{table: imagenet-compression}. Unlike the experiments performed previously, we start with a pre-trained model and then perform architecture learning (AL) on the learnt weights. We see that our method performs almost as well as the state of the art compression methods. This means that one can simply use a smaller neural network instead of using weight re-parameterization techniques (FastFood, SVD) on a large network.

Further, many compression methods are formulated keeping only fully-connected layers in mind. For tasks like Semantic Segmentation, networks with only convolutional layers are used \cite{long2015fully}. Our results show that the proposed method can successfully prune both fully connected neurons and convolutional filters. Further, ours (along with SVD) is among the \textit{few} compression methods that can utilize dense matrix computations, whereas all other methods require specialized kernels for sparse matrix computations \cite{han2015learning} or custom implementations for diagonal matrix multiplication \cite{yang2014deep},  etc.

\begin{table*}[htbp]
\centering
\begin{tabular}{c|c|c|c}
\hline 
\textbf{Method} & \textbf{Params} & \textbf{Accuracy (\%)} & \textbf{Compression (\% )} \\  
\hline \hline
Reference Model (CaffeNet) & 60.9M & 57.41 & 0\\
Neuron Pruning (\cite{BMVC2015_31}) & 39.6M & 55.60 & 35\\
SVD-quarter-F (\cite{yang2014deep}) & 25.6M & 56.19 & 58\\
Adaptive FastFood 16 (\cite{yang2014deep}) & 18.7M & 57.10 & 69\\
\hline \hline

AL-conv-fc & 19.6M & 55.90 & 68\\
AL-fc & 19.8M & 54.30 & 68\\
AL-conv & 47.8M & 55.87 & 22\\

\hline
\end{tabular} 
\vspace{5pt}
\caption{\small{Compression performance on CaffeNet.}}
\label{table: imagenet-compression}
\vspace{-0.5cm}
\end{table*}

\begin{table*}[htbp]
\centering
\begin{tabular}{c|c|c}
\hline 
\textbf{Method} & \textbf{Layers Learnt} & \textbf{Architecture}  \\  
\hline \hline
Baseline & N/A & 96~~256~~384~~384~~256~~4096~~4096~~1000  \\
AL-fc & fc[6,7] & 96~~256~~384~~384~~256~~\textcolor{red}{1536}~~\textcolor{red}{1317}~~1000  \\
AL-conv & conv[1,2,3,4,5] & \textcolor{red}{80}~~\textcolor{red}{127}~~\textcolor{red}{264}~~\textcolor{red}{274}~~\textcolor{red}{183}~~4096~~4096~~1000  \\
AL-conv-fc & conv[5] - fc[6,7] & 96~~256~~384~~384~~\textcolor{red}{237}~~\textcolor{red}{1761}~~\textcolor{red}{1661}~~1000  \\
\hline 
\end{tabular} 
\vspace{5pt}
\caption{\small{Architectures learnt by our method whose performance is given in Table \ref{table: imagenet-compression}.}}
\label{table: imagenet-arch}
\vspace{-0.5cm}
\end{table*}

\vspace{-0.25cm}
\section{Conclusions}
\vspace{-0.25cm}
We have presented a method to learn a neural network's architecture along with weights. Rather than directly selecting width and depth of networks, we introduced a small number of real-valued hyper-parameters which selected width and depth for us. We also saw that we get smaller architectures for MNIST and ImageNet datasets that perform on par with the large architectures. Our method is very simple and straightforward, and can be suitably applied to any neural network. This can also be used as a tool to further explore the dependence of architecture on the optimization and convergence of neural networks. 


\bibliography{refs_icml}

\begin{thebibliography}{26}
\providecommand{\natexlab}[1]{#1}
\providecommand{\url}[1]{\texttt{#1}}
\expandafter\ifx\csname urlstyle\endcsname\relax
  \providecommand{\doi}[1]{doi: #1}\else
  \providecommand{\doi}{doi: \begingroup \urlstyle{rm}\Url}\fi

\bibitem[Bergstra et~al.(2010)Bergstra, Breuleux, Bastien, Lamblin, Pascanu,
  Desjardins, Turian, Warde-Farley, and Bengio]{bergstra2010theano}
James Bergstra, Olivier Breuleux, Fr{\'e}d{\'e}ric Bastien, Pascal Lamblin,
  Razvan Pascanu, Guillaume Desjardins, Joseph Turian, David Warde-Farley, and
  Yoshua Bengio.
\newblock Theano: a cpu and gpu math expression compiler.
\newblock In \emph{Proceedings of the Python for scientific computing
  conference (SciPy)}, volume~4, page~3. Austin, TX, 2010.

\bibitem[Collins and Kohli(2014)]{DBLP:journals/corr/CollinsK14}
Maxwell~D. Collins and Pushmeet Kohli.
\newblock Memory bounded deep convolutional networks.
\newblock \emph{CoRR}, abs/1412.1442, 2014.
\newblock URL \url{http://arxiv.org/abs/1412.1442}.

\bibitem[Denil et~al.(2013)Denil, Shakibi, Dinh, de~Freitas,
  et~al.]{denil2013predicting}
Misha Denil, Babak Shakibi, Laurent Dinh, Nando de~Freitas, et~al.
\newblock Predicting parameters in deep learning.
\newblock In \emph{Advances in Neural Information Processing Systems}, pages
  2148--2156, 2013.

\bibitem[Denton et~al.(2014)Denton, Zaremba, Bruna, LeCun, and
  Fergus]{denton2014exploiting}
Emily~L Denton, Wojciech Zaremba, Joan Bruna, Yann LeCun, and Rob Fergus.
\newblock Exploiting linear structure within convolutional networks for
  efficient evaluation.
\newblock In \emph{Advances in Neural Information Processing Systems}, pages
  1269--1277, 2014.

\bibitem[Fahlman and Lebiere(1989)]{fahlman1989cascade}
Scott~E Fahlman and Christian Lebiere.
\newblock The cascade-correlation learning architecture.
\newblock 1989.

\bibitem[Gong et~al.(2014)Gong, Liu, Yang, and Bourdev]{gong2014compressing}
Yunchao Gong, Liu Liu, Ming Yang, and Lubomir Bourdev.
\newblock Compressing deep convolutional networks using vector quantization.
\newblock \emph{arXiv preprint arXiv:1412.6115}, 2014.

\bibitem[Han et~al.(2015)Han, Pool, Tran, and Dally]{han2015learning}
Song Han, Jeff Pool, John Tran, and William~J Dally.
\newblock Learning both weights and connections for efficient neural networks.
\newblock \emph{arXiv preprint arXiv:1506.02626}, 2015.

\bibitem[Hassibi et~al.(1993)Hassibi, Stork, et~al.]{hassibi1993second}
Babak Hassibi, David~G Stork, et~al.
\newblock Second order derivatives for network pruning: Optimal brain surgeon.
\newblock \emph{Advances in Neural Information Processing Systems}, pages
  164--164, 1993.

\bibitem[He et~al.(2015)He, Zhang, Ren, and Sun]{he2015delving}
Kaiming He, Xiangyu Zhang, Shaoqing Ren, and Jian Sun.
\newblock Delving deep into rectifiers: Surpassing human-level performance on
  imagenet classification.
\newblock \emph{arXiv preprint arXiv:1502.01852}, 2015.

\bibitem[Jaderberg et~al.(2014)Jaderberg, Vedaldi, and
  Zisserman]{jaderberg2014speeding}
Max Jaderberg, Andrea Vedaldi, and Andrew Zisserman.
\newblock Speeding up convolutional neural networks with low rank expansions.
\newblock In \emph{Proceedings of the British Machine Vision Conference}. BMVA
  Press, 2014.

\bibitem[Jia et~al.(2014)Jia, Shelhamer, Donahue, Karayev, Long, Girshick,
  Guadarrama, and Darrell]{jia2014caffe}
Yangqing Jia, Evan Shelhamer, Jeff Donahue, Sergey Karayev, Jonathan Long, Ross
  Girshick, Sergio Guadarrama, and Trevor Darrell.
\newblock Caffe: Convolutional architecture for fast feature embedding.
\newblock \emph{arXiv preprint arXiv:1408.5093}, 2014.

\bibitem[Krizhevsky et~al.(2012)Krizhevsky, Sutskever, and
  Hinton]{krizhevsky2012imagenet}
Alex Krizhevsky, Ilya Sutskever, and Geoffrey~E Hinton.
\newblock Imagenet classification with deep convolutional neural networks.
\newblock In \emph{Advances in Neural Information Processing Systems}, pages
  1097--1105, 2012.

\bibitem[Kulkarni et~al.(2015)Kulkarni, Zepeda, Jurie, Pérez, and
  Chevallier]{BMVC2015_23}
Praveen Kulkarni, Joaquin Zepeda, Frederic Jurie, Patrick Pérez, and Louis
  Chevallier.
\newblock Learning the structure of deep architectures using l1 regularization.
\newblock In Mark W.~Jones Xianghua~Xie and Gary K.~L. Tam, editors,
  \emph{Proceedings of the British Machine Vision Conference (BMVC)}, pages
  23.1--23.11. BMVA Press, September 2015.
\newblock ISBN 1-901725-53-7.
\newblock \doi{10.5244/C.29.23}.
\newblock URL \url{https://dx.doi.org/10.5244/C.29.23}.

\bibitem[LeCun et~al.(1989)LeCun, Denker, Solla, Howard, and
  Jackel]{lecun1989optimal}
Yann LeCun, John~S Denker, Sara~A Solla, Richard~E Howard, and Lawrence~D
  Jackel.
\newblock Optimal brain damage.
\newblock In \emph{Advances in Neural Information Processing Systems},
  volume~2, pages 598--605, 1989.

\bibitem[LeCun et~al.(1998)LeCun, Bottou, Bengio, and
  Haffner]{lecun1998gradient}
Yann LeCun, L{\'e}on Bottou, Yoshua Bengio, and Patrick Haffner.
\newblock Gradient-based learning applied to document recognition.
\newblock \emph{Proceedings of the IEEE}, 86\penalty0 (11):\penalty0
  2278--2324, 1998.

\bibitem[Long et~al.(2015)Long, Shelhamer, and Darrell]{long2015fully}
Jonathan Long, Evan Shelhamer, and Trevor Darrell.
\newblock Fully convolutional networks for semantic segmentation.
\newblock In \emph{Proceedings of the IEEE Conference on Computer Vision and
  Pattern Recognition}, pages 3431--3440, 2015.

\bibitem[Murray and Ng(2010)]{murray2010algorithm}
Walter Murray and Kien-Ming Ng.
\newblock An algorithm for nonlinear optimization problems with binary
  variables.
\newblock \emph{Computational Optimization and Applications}, 47\penalty0
  (2):\penalty0 257--288, 2010.

\bibitem[Russakovsky et~al.(2015)Russakovsky, Deng, Su, Krause, Satheesh, Ma,
  Huang, Karpathy, Khosla, Bernstein, Berg, and Fei-Fei]{ILSVRC15}
Olga Russakovsky, Jia Deng, Hao Su, Jonathan Krause, Sanjeev Satheesh, Sean Ma,
  Zhiheng Huang, Andrej Karpathy, Aditya Khosla, Michael Bernstein,
  Alexander~C. Berg, and Li~Fei-Fei.
\newblock {ImageNet Large Scale Visual Recognition Challenge}.
\newblock \emph{International Journal of Computer Vision (IJCV)}, 2015.
\newblock \doi{10.1007/s11263-015-0816-y}.

\bibitem[Simonyan and Zisserman(2015)]{Simonyan15}
K.~Simonyan and A.~Zisserman.
\newblock Very deep convolutional networks for large-scale image recognition.
\newblock In \emph{International Conference on Learning Representations}, 2015.

\bibitem[Snoek et~al.(2012)Snoek, Larochelle, and Adams]{snoek2012practical}
Jasper Snoek, Hugo Larochelle, and Ryan~P Adams.
\newblock Practical bayesian optimization of machine learning algorithms.
\newblock In \emph{Advances in neural information processing systems}, pages
  2951--2959, 2012.

\bibitem[Srinivas and Babu(2015)]{BMVC2015_31}
Suraj Srinivas and R.~Venkatesh Babu.
\newblock Data-free parameter pruning for deep neural networks.
\newblock In Mark W.~Jones Xianghua~Xie and Gary K.~L. Tam, editors,
  \emph{Proceedings of the British Machine Vision Conference (BMVC)}, pages
  31.1--31.12. BMVA Press, September 2015.
\newblock ISBN 1-901725-53-7.
\newblock \doi{10.5244/C.29.31}.
\newblock URL \url{https://dx.doi.org/10.5244/C.29.31}.

\bibitem[Stanley and Miikkulainen(2002)]{stanley2002evolving}
Kenneth~O Stanley and Risto Miikkulainen.
\newblock Evolving neural networks through augmenting topologies.
\newblock \emph{Evolutionary computation}, 10\penalty0 (2):\penalty0 99--127,
  2002.

\bibitem[Stanley et~al.(2009)Stanley, D'Ambrosio, and
  Gauci]{stanley2009hypercube}
Kenneth~O Stanley, David~B D'Ambrosio, and Jason Gauci.
\newblock A hypercube-based encoding for evolving large-scale neural networks.
\newblock \emph{Artificial life}, 15\penalty0 (2):\penalty0 185--212, 2009.

\bibitem[Szegedy et~al.(2015)Szegedy, Liu, Jia, Sermanet, Reed, Anguelov,
  Erhan, Vanhoucke, and Rabinovich]{Szegedy_2015_CVPR}
Christian Szegedy, Wei Liu, Yangqing Jia, Pierre Sermanet, Scott Reed, Dragomir
  Anguelov, Dumitru Erhan, Vincent Vanhoucke, and Andrew Rabinovich.
\newblock Going deeper with convolutions.
\newblock In \emph{The IEEE Conference on Computer Vision and Pattern
  Recognition (CVPR)}, June 2015.

\bibitem[Yang et~al.(2014)Yang, Moczulski, Denil, de~Freitas, Smola, Song, and
  Wang]{yang2014deep}
Zichao Yang, Marcin Moczulski, Misha Denil, Nando de~Freitas, Alex Smola,
  Le~Song, and Ziyu Wang.
\newblock Deep fried convnets.
\newblock \emph{arXiv preprint arXiv:1412.7149}, 2014.

\bibitem[Yao(1999)]{yao1999evolving}
Xin Yao.
\newblock Evolving artificial neural networks.
\newblock \emph{Proceedings of the IEEE}, 87\penalty0 (9):\penalty0 1423--1447,
  1999.

\end{thebibliography}

\part*{Supplementary Material}

\section*{Properties of the method}
Here we identify a few properties of our architecture selection method.

\begin{enumerate}
\item \textbf{Non-redundancy of architecture}: The learnt final architecture must not have any redundant neurons. Removing neurons should necessarily degrade performance.
\item \textbf{Local-optimality of weights}: The performance of the learnt final architecture must at least be equal to a trained neural network initialized with this final architecture. 
\item \textbf{Mirroring data-complexity}: A `harder' dataset should result in a larger model than an `easier' dataset.
\end{enumerate}

We intuitively observe that all these properties would automatically hold if a `master' property which requires both the architecture and the weights be globally optimal holds. Given that the optimization objective of neural networks is highly non-convex, global optimality cannot be guaranteed. As a result, we restrict ourselves to studying the three properties listed.

In the text that follows, we provide statements that hold for our method. These are obtained by analysing widths of each layer of a neural network assuming that depth is never collapsed. In other words, these hold for neural networks with a single hidden layer. Proofs are provided in a later section.

\subsection*{Non-redundancy of architecture}
This is an important property that forms the main motivation for doing architecture-learning. Such a procedure can replace the node-pruning techniques that are used to compress neural networks. 

\begin{prop}
At convergence, the loss ($\ell$) of the proposed method over the train set satisfies $\dfrac{\partial \ell}{\partial \Phi} < 0$
\label{prop: nonred}
\end{prop}

This statement implies that change in architecture is inversely proportional to change in loss. In other words, if the architecture grows smaller, the loss must increase. While there isn't a strict relationship between loss and accuracy, a high loss generally indicates worse accuracy.

\subsection*{Local Optimality of weights}
The proposed method learns both architecture and weights. What would happen if we initialized a neural network with this learnt architecture, and proceeded to learn only the weights? This property ensures that in both cases we fall into a local minimum with architecture $\Phi$.

\begin{prop}
Let $\ell_1$ be the loss over the train set at convergence obtained by training a neural network on data $\mathcal{D}$ with a fixed architecture $\Phi$. Let $\ell_2$ be the loss at convergence when the neural network is trained with the proposed method on data $\mathcal{D}$ such that it results in the same final architecture $\Phi$. Then, $\dfrac{\partial \ell_1}{\partial \theta} < \epsilon$ and $\dfrac{\partial \ell_2}{\partial \theta} < \epsilon$ for any $\epsilon \rightarrow 0$.
\label{prop:localopt}
\end{prop}

\subsection*{Mirroring data-complexity}
Characterizing data-complexity has traditionally been hard. Here, we consider the following approach.

\begin{prop}
Let $\mathcal{D}_1$ and $\mathcal{D}_2$ be two datasets which produce train losses $\ell_1$ and $\ell_2$ upon training with a fixed architecture $\Phi$ such that $\ell_1 > \ell_2$.
When trained with the proposed method, the final architectures $\hat{\Phi}_1$ and $\hat{\Phi}_2$ (corresponding to  $\mathcal{D}_1$ and $\mathcal{D}_2$) satisfy the relation $\| \hat{\Phi}_1 \| > \| \hat{\Phi}_2 \|$ at convergence.
\label{prop: mirror}
\end{prop}

Here, $\mathcal{D}_1$ is the `harder' dataset because it produces a higher loss on the same neural network architecture. As a result, the `harder' dataset always produces a larger final architecture. We do not provide a proof for this statement. Instead, we experimentally verify this in Section 4.2.

\section*{Proofs of Propositions}

Let $E = \ell + \lambda_b \mathcal{R}_b + \lambda_m \mathcal{R}_m$ be total objective function, where $\mathcal{R}_b$ is the binarizing regularizer, $\mathcal{R}_m = \| \phi \|$ is the model complexity term. At convergence, we assume that $\mathcal{R}_b = 0$ as the corresponding weights are all binary or close to binary. Let the maximum step size (due to gradient clipping) for \textbf{w} and \textbf{d} be $s$.
 
\begin{proof}[Proof of proposition 1] At convergence, we assume \\$\dfrac{\partial E}{\partial \phi} < \epsilon$, for some $\epsilon \rightarrow 0_+$.

\begin{equation*}
\dfrac{\partial \ell}{\partial \phi} < - \lambda_m \dfrac{\partial \| \phi \|}{\partial \phi} +  \epsilon 
\implies \dfrac{\partial \ell}{\partial \phi} < - \lambda_m + \epsilon 
\implies \dfrac{\partial \ell}{\partial \phi} < 0 
\end{equation*}

for some $\epsilon$ sufficiently small.

\end{proof}

\begin{proof}[Proof of proposition 2] Let $\mathcal{R}_b = 0$ at $t_1^{th}$ iteration with architecture $\Phi_1$. Let $\Phi_2$ be the architecture at iteration $t_2 > t_1$ such that at iterations $ t_1 < t < t_2$, architecture is $\Phi_1$.

$\implies \exists$ an iteration $ t_1 < t < t_2$ such that $\mathcal{R}_b > s (1 - s) = s_1$, $s$ being the maximum step size. 

Let $q = \dfrac{\partial \ell_2}{\partial \theta_2}$. Let $\lambda_b$ be parameterized by $k$ as follows.

\begin{align*}
\centering
\lambda_b ~s_1= \mathbbm{E}_{\mathcal{D}} (q) + k \sigma ~~ & \mathrm{where} ~~ \sigma = \mathbbm{E}_{\mathcal{D}} (q - \mathbbm{E}_{\mathcal{D}}(q))^2 \\
\centering
\mathrm{If}~~ k \rightarrow \infty ~~ & \mathrm{then}~~ \mathbbm{P} (q > \lambda_b ~ s_1) \rightarrow 0
\end{align*}

Hence, for large enough $\lambda_b$, $\Phi_2 = \Phi_1$. After $T >> t$ iterations, we have 

\begin{equation}
\dfrac{\partial \ell_1}{\partial \theta_1} < \epsilon ~~\mathrm{and}~~ \dfrac{\partial \ell_2}{\partial \theta_2} < \epsilon
\label{eqn:pde}
\end{equation}

for some $\epsilon \rightarrow 0_+$. However, if $\theta_1 \in \mathbbm{R}^{d_1}$, then $\theta_2 \in \mathbbm{R}^{d_2}$, such that $d_1 < d_2$.

Without loss of generality, let us assume that neurons corresponding to first $d_1$ weights are selected for, while the rest are inactive. As a result,  $\dfrac{\partial \ell_2}{\partial \theta_2 (d)} = 0$, for $d \in [d_1,d_2]$. Hence, the following holds $\dfrac{\partial \ell_2}{\partial \theta_1} < \epsilon$. This, along with equation \ref{eqn:pde}, proves the assertion.

\end{proof}

\section*{Hyper-parameter selection}
For effective usage of our method, we need a good set of $\lambda$s. Here, we describe how to do so practically.

First, we set $\lambda_3$ to a low value based on the initial widths and loss values. Recall that this value multiplies with the number of neurons in the cost function. That is, if a network has a layer with $n$ neurons, we get $\lambda_3 \times n$. Hence, if $n$ multiplies by 10, $\lambda_3$ divides by 10, so that the regularizer value remains the same. We used $\lambda_3 = 10^{-5}$ for MNIST-network and $\lambda_3 = 10^{-6}$ for AlexNet. For a given initial architecture, a large $\lambda_3$ places more emphasis on getting small models than reducing loss.

Second, we set $\lambda_1$ to be about $\sim 2$ times $\lambda_3$. Using a positive $\lambda_3$ shifts the curve in Fig. \ref{fig:NN}\textcolor{red}{(b)} to the right. By letting $\lambda_3 = \lambda_1$, the curve shifts to the extreme right with the peak at $x = 1$. Hence if $\lambda_3 = k \times \lambda_1$, we set $ 0 < k < 1$.

We simply set $\lambda_2$ and $\lambda_4$ to $1/10 ^{th}$ of $\lambda_1$ and $\lambda_3$ respectively.

\end{document}